\documentclass{article}

\usepackage{techreport}

\usepackage{graphicx}
\usepackage{booktabs}
\usepackage{array}
\usepackage{amsmath}
\usepackage{amssymb}
\usepackage{amsfonts}
\usepackage{multirow}
\usepackage{verbatim}
\usepackage{caption}
\usepackage{longtable}
\usepackage{supertabular}
\usepackage{float}
\usepackage{enumitem}
\usepackage{tablefootnote}
\usepackage[round,semicolon]{natbib}
\usepackage{xcolor}
\usepackage{colortbl}
\usepackage{xspace}
\usepackage{textcomp}
\usepackage{makecell}

\usepackage{lscape} 
\usepackage{siunitx}

\usepackage{pifont}%
\usepackage{scrextend}

\usepackage{array}
\usepackage{tgpagella}
\usepackage{latexsym}
\usepackage[T1]{fontenc}
\usepackage[utf8]{inputenc}
\usepackage{microtype}
\definecolor{mydarkblue}{rgb}{0,0.08,0.45}
\definecolor{mylightblue}{RGB}{39,114,191}
\usepackage[colorlinks,citecolor=mylightblue,urlcolor=mylightblue,linkcolor=mylightblue]{hyperref}
\usepackage{url}            %
\usepackage{nicefrac}       %
\usepackage{changepage}
\usepackage{xargs}          %
\usepackage{wrapfig,lipsum,booktabs}
\usepackage{longtable}
\usepackage{subcaption}
\usepackage{endnotes}

\usepackage{pgfplots}
\usetikzlibrary{pgfplots.groupplots}
\pgfplotsset{compat=1.3}

\usepackage[most]{tcolorbox}

\usepackage[capitalize,noabbrev]{cleveref}

\usepackage{multicol}
\usepackage{multirow}

\usepackage{tcolorbox}
\usepackage{titlesec}
\titleformat*{\section}{\large\bfseries}

\definecolor{myyellowgreen}{RGB}{154, 205, 50}

\usepackage{minitoc}

\newcommand{\IGNORE}[1]{}
\usepackage{algorithm}
\usepackage{algorithmicx}
\usepackage{algpseudocode}

\newcommand\simplefootnote[1]{%
  \begingroup
  \renewcommand\thefootnote{}\footnote{#1}%
  \addtocounter{footnote}{-1}%
  \endgroup
}

\title{\vspace{-2em}%
  \hrule height 4pt%
  \vskip 0.25in%
  \vskip -\parskip%
  \textbf{
FP8-LM: Training FP8 Large Language Models
}
  \vskip 0.2in%
  \vskip -\parskip%
  \hrule height 1pt%
  \vskip 0.09in}

\author{
Houwen Peng $^*$ \hspace{0.25cm} 
Kan Wu $^*$ \hspace{0.25cm} 
Yixuan Wei $^*$  \vspace{0.07cm} \and
Guoshuai Zhao \hspace{0.12cm}
Yuxiang Yang \hspace{0.12cm}
Ze Liu \hspace{0.12cm}
Yifan Xiong \hspace{0.12cm}
Ziyue Yang \vspace{0.07cm} \and
Bolin Ni \hspace{0.12cm}
Jingcheng Hu \hspace{0.12cm}
Ruihang Li \hspace{0.12cm}
Miaosen Zhang \hspace{0.12cm}
Chen Li \hspace{0.12cm}
Jia Ning \hspace{0.12cm}
Ruizhe Wang \hspace{0.12cm}
Zheng Zhang \vspace{0.07cm} \and
Shuguang Liu \hspace{0.12cm}
Joe Chau \hspace{0.12cm}
Han Hu $^\dagger$ \hspace{0.12cm}
Peng Cheng $^\dagger$ \hspace{0.12cm}
\vspace{0.8cm} \\
\textbf{\normalsize Microsoft Azure and Microsoft Research}
}

\date{}

\begin{document}
\maketitle

\simplefootnote{Contributions for all the authors can be found in Section \ref{contribution}.}
\simplefootnote{* equal work \hspace{0.12cm} $^\dagger$ contact: \{hanhu | pengc\}@microsoft.com}

\vspace{-1em}
\begin{abstract}
In this paper, we explore FP8 low-bit data formats for efficient training of large language models (LLMs). Our key insight is that most variables, such as gradients and optimizer states, in LLM training can employ low-precision data formats without compromising model accuracy and requiring no changes to hyper-parameters. Specifically, we propose a new FP8 automatic mixed-precision framework for training LLMs. This framework offers three levels of FP8 utilization to streamline mixed-precision and distributed parallel training for LLMs. It gradually incorporates 8-bit gradients, optimizer states, and distributed learning in an incremental manner. Experiment results show that, during the training of GPT-175B model on  H100 GPU platform, our FP8 mixed-precision training framework not only achieved a remarkable 39\% reduction in real memory usage but also ran 75\% faster than the widely adopted BF16 framework (i.e., Megatron-LM), surpassing the speed of Nvidia Transformer Engine by 37\%. 
This largely reduces the training costs for large foundation models. Furthermore, our FP8 mixed-precision training methodology is generic. It can be seamlessly applied to other tasks such as LLM instruction tuning and reinforcement learning with human feedback, offering savings in fine-tuning expenses. Our FP8 low-precision training framework is open-sourced at \href{https://github.com/Azure/MS-AMP}{aka.ms/MS.AMP}.
\end{abstract}

\section{Introduction}

Large language models (LLMs) \citep{gpt3, megatron-nlg, palm, opt} have demonstrated unprecedented capabilities in language comprehension and generation, leading to breakthroughs in reasoning, math, science, and many other tasks \citep{gpt4, palm2}. %
However, training LLMs is extremely costly. For example, PaLM takes 6,144 TPUv4 chips to train a 540B model, while GPT-3 175B consumes several thousand petaflop/s-days of compute for pre-training \citep{palm, gpt3}. This motivates the needs of reducing the training costs of LLMs, especially for the scaling of next-generation super-intelligent models. %

Low-precision training is one of the most promising directions to reduce the costs, as it can provide high speed, small memory footprint, and low communication overhead. Most existing training systems, \emph{e.g.}, Megatron-LM \citep{megatron-lm}, MetaSeq \citep{opt}, and Colossal-AI \citep{colossal-ai}, train LLMs with either FP32 full-precision or FP16/BF16 mixed-precision by default. This is not essential, however, to achieve full accuracy for large models. With the release of Nvidia H100 GPU, FP8 is becoming the next-generation datatype for low-precision representation \citep{h100-whitepaper, fp8-dl}. Theoretically, FP8 can achieve $2\times$ speed-up, 50\% - 75\% memory cost savings, and 50\% - 75\% communication savings compared with current 16-bit and 32-bit floating point mixed-precision training, which is very promising for scaling-up next-generation foundation models.

Unfortunately, the current support for FP8 training is rare and limited. The only usable framework is the Nvidia Transformer Engine (TE) \citep{te}, but it applies FP8 solely for GEMM computation and still retains master weights and gradients using high precision, \emph{e.g.}, FP16 or FP32.
As a result, the end-to-end speed-up, memory and communication cost savings are very limited, which does not fully unveil the power of FP8. To address this issue, we propose an extremely optimized FP8 mixed-precision framework for LLM training. The core idea is to infiltrate FP8 compute, storage, and communication into the whole progress of large model training, making the forward and backward pass all used the low-precision FP8, thus largely reducing system workloads compared to previous frameworks \citep{amp, te, fp8-dl}.
Specifically, we design three optimization levels that utilize FP8 to streamline mixed-precision and distributed training. The three levels gradually incorporate 8-bit collective communication, optimizer, and distributed parallel training in an incremental manner. The higher optimization level indicates using more FP8 during LLM training. Moreover, for large-scale training, such as GPT-175B trained on thousand of GPUs, our framework provides FP8 low-bit parallelism, including tensor, pipeline, and sequence parallelism, paving the way to next-generation low-precision parallel training. 

\begin{figure}[t]
\centering
\includegraphics[height=5.5cm]{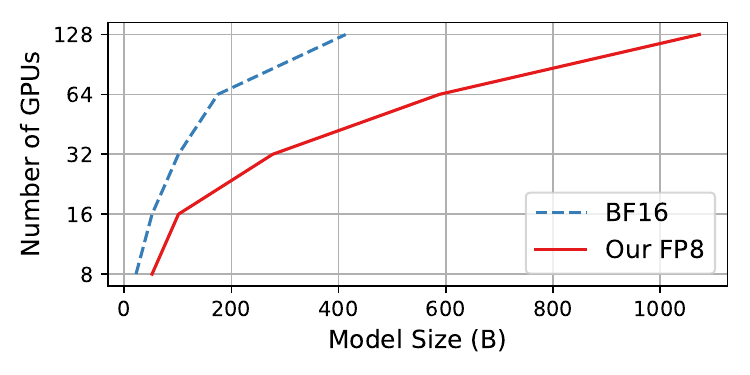}
\caption{An analysis of comparing the maximum model sizes attainable  through the utilization of either the prevalent BF16 or our FP8 mixed-precision training approach on a cluster of Nvidia H100 GPUs with 80GB memory.}
\label{fig.gpt}
\vspace{-5mm}
\end{figure}

Training LLMs with FP8 is non-trivial. The challenges stem from issues such as data underflow or overflow, coupled with quantization errors arising from the narrower dynamic range and reduced precision inherent in FP8 data formats. These challenges cause numerical instabilities and irreversible divergences throughout the training process. %
To tackle them, we propose two techniques:  \emph{precision decoupling} and \emph{automatic scaling} for preventing the loss of critical information. The former one involves decoupling the influence of data precision on parameters such as weights, gradients, optimizer states, and assigning reduced precision to components that are not precision sensitive. The latter one is to preserve gradient values within the representation range of FP8 data formats  through the dynamic adjustment of tensor scaling factors, thereby alleviating underflow and overflow occurrences during all-reduce communication. %

To validate the proposed FP8 low-precision framework, we apply it to GPT-style model training, encompassing both pre-training and supervised fine-tuning (SFT).  %
The experimental results demonstrate the effectiveness of our FP8 methodology, yielding substantial benefits including a 29\% to 39\% reduction in real memory usage (\emph{e.g.}, 29\% reduction for GPT-7B while 39\% for GPT-175B ) and a notable  63\% to 65\% decrease in weight-related communication overhead compared to the prevalent BF16 mixed-precision training approach. Without changes to any hyper-parameters, such as learning rate and weight decay, the models trained using FP8 exhibit  performance equivalency to those employing BF16 high precision, both in pre-training and downstream tasks. It is noteworthy that during the training of GPT-175B model, our FP8 mix-precision framework reduces training time by 37\% compared to TE \citep{te},  while consuming 42\% less memory on H100 GPU platform. 
More importantly, %
the reduction in costs achieved through the utilization of low-precision FP8 can be further increased, as the scale of models continues to expand, which is presented in Fig.~\ref{fig.gpt}.

For fine-tuning, we employ FP8 mixed-precision for instruction tuning and reinforcement learning with human feedback (RLHF) to better align pre-trained LLMs with end tasks and user preferences. %
Specifically, we fine-tune pre-trained models on publicly user-shared instruction-following data \citep{sharegpt}. The models tuned with our FP8 mixed-precision demonstrate comparable performance to those utilizing the half-precision BF16 \citep{mt-bench} on the AlpacaEval \citep{alpaca_eval} and MT-Bench \citep{mt-bench} benchmarks, while achieving 27\% improvements in training speed. Moreover, FP8 mixed-precision exhibits considerable potentials in RLHF, a process that necessitates loading multiple models during training. %
Through the utilization of FP8 in training, the prevalent RLHF framework AlpacaFarm \citep{alpacafarm} can  yield a 32\% reduction in model weights and a 62\% reduction in optimizer states' memory consumption. 
This further demonstrates the versatility and adaptability of our FP8 low-precision training framework.

We are making the following contributions to drive the design of next-generation FP8 low-precision training\IGNORE{ and inference framework} for LLMs.

\begin{itemize}

 \item  A new FP8 mixed-precision training framework. It unlocks 8-bit weights, gradients, optimizer, and distributed training gradually in an add-on fashion, which is convenient in use. %
 This 8-bit framework can be used as a simple drop-in replacement for existing 16/32-bit mixed-precision counterparts, without requiring any changes to the hyper-parameters and training receipts.  Additionally, we provide a Pytorch implementation that enables 8-bit low-precision training in a few lines of code. %

\item A new family of GPT-style models trained with FP8. We apply the proposed FP8 scheme to GPT pre-training and fine-tuning (\emph{i.e.}, SFT and RLHF), and demonstrate its potentials on a variety of model scales ranging from 7B to 175B parameters. We equip prevalent parallel computation paradigms with FP8 supports, including tensor, pipeline, and sequence parallelisms, enabling the utilization of FP8 to train large foundation models. We open-source the first FP8 GPT training codebase based upon Megatron-LM \citep{megatron-lm} implementation. %
\end{itemize}

We expect the release of our FP8 framework will establish a new paradigm for next-generation low-precision training system dedicated to large foundation models.%

\section{FP8 LLMs}

Mixed-precision \citep{amp} has been widely used in LLM training to improve compute and memory efficiency. The most popular mixed-precision schemes are FP16-FP32 and BF16-FP32. Because of the restricted numerical range of FP16, FP16-FP32 scheme has been known instabilities for training large models  \citep{gopher,glm-130b}. Consequently, the community now commonly adopts BF16-FP32 for training LLMs, such as Megatron-Turing NLG-530B \citep{megatron-nlg}, Bloom-175B \citep{bloom} and Gopher \citep{gopher}. The underlying reason is that BF16 has a wide dynamic range to maintain numerical stability while matching the performance of the full-precision FP32. Moreover, BF16 employs half the number of bits as compared to FP32, thus reducing considerable memory footprints while improving compute efficiency. 

FP8 is a natural evolution from 16-bit data formats to further reducing computing costs. However, training LLMs with reduced-precision FP8 poses new challenges. The dynamic range and representation precision of FP8\footnote{The details of FP8 data formats are presented in Appendix~\ref{appendix.A}.} are much lower than BF16 and FP16, which inevitably induces more training collapses, such as loss spikes or even NaNs. To address the issues, tensor scaling techniques are proposed \citep{sunxiao-fp8, fp8-dl}. The core idea is multiplying higher precision values with a scaling factor prior to their casting to FP8 in order to move them into a range that better overlaps with the representable range of a corresponding FP8 format\footnote{The details of FP8 tensor scaling are introduced in Appendix \ref{appendix.B}.} 
\citep{fp8-dl}. Such a per-tensor scaling technique reduces data quantization errors while improving numerical stability and accuracy, thus enabling the utilization of the lower-precision FP8 for training large models.

Unfortunately, the current support for FP8 low-precision training is restricted. Nvidia TE \citep{te} only supports FP8 compute for linear layers in Transformer \citep{transformer}, while leaving all other operations, such as weight update and gradient synchronization, still using higher precision. In this work, we present an extremely optimized FP8 mixed-precision strategy for LLM training. The new FP8 optimization includes three key perspectives: FP8 communication, FP8 optimizer, and FP8 distributed training. By integrating these aspects, the training of LLMs such as the 175B GPT-3 model can fully harness the advantages of FP8 low-precision and improve training efficiency.

\subsection{FP8 Gradient and All-Reduce Communication}
\label{sec.fp8grad}
Existing mixed-precision training methodologies \citep{amp, te} typically employ 16-bit or 32-bit datatype for the computation and storage of gradients, resulting in a high bandwidth requirement for collective communication throughout the training process. We found that directly applying FP8 to gradients leads to a decrease in accuracy. %
The fundamental issue lies in the underflow and overflow problems arising from the low-bit all-reduce operation. 
Specifically, there are two standard methods aggregating gradients across GPUs during all-reduce: \emph{pre-scaling} and \emph{post-scaling}. Pre-scaling divides the gradient $g_i$ calculated on the $i$-th GPU by the total number of GPUs (\emph{i.e.}, $N$) before being summed, which is formulated as:
\begin{equation}
g = g_{1}/N + g_{2}/N + \cdots +g_{N}/N.
\end{equation}
When $N$ is large, this division can cause data underflow, especially for FP8 low-precision representation of gradients. To mitigate this issue, post-scaling performs the gradient summation first, followed by the division scaling during the gradient collection process: 
\begin{equation}
g = (g_{1} + g_{2} + \cdots + g_{N})/N.
\end{equation}
This post-scaling approach keeps the gradients close to the maximum value of the FP8 datatype, effectively alleviating the underflow issue. However, this approach  encounters overflow issues when aggregating gradients.

In contrast, we propose an \emph{automatic scaling} technique to resolve both the underflow and overflow issues in the pre-scaling and post-scaling approaches. To be specific, we introduce an auto-scaling factor $\mu$, that changes on the fly during the training, %
to reduce the occurrences of overflow and underflow in gradients:
\begin{equation}
g'_{i} = \mu \cdot g_{i}.
\label{eq_mu}
\end{equation}
A statistical analysis is conducted on the gradient values of $g'_{i}$,  with the objective of quantifying the proportion of values that attains the maximum feasible value within the FP8 representation range. If the ratio of the maximum value exceeds a specified threshold, \emph{i.e.}, 0.001\%, $\mu$ is set to $1/2$ in the subsequent training step, thereby mitigating the risk of overflow. Conversely, when the ratio consistently remains the threshold, we opt to exponentially increase $\mu$ to $2$ over  the span of 1,000 training steps, thereby effectively mitigating the risk of underflow occurrences.

Another key obstacle of FP8 collective communication lies in devising an effective strategy to manage the tensor-wise scaling factors that are associated with each gradient tensor. The current NCCL implementation \citep{nccl} lacks the capability of performing all-reduce operation considering the additional tensor-wise scaling factors. Meanwhile, efficient implementation is also very challenging, especially considering that the NCCL gradient summation operates at sub-tensor level. This complexity increases significantly when incorporating updates for tensor-wise scaling factors. 
To overcome this issue, we propose a new mechanism that scales FP8 gradients across GPUs using a single shared scalar. To be specific, let $(g'_i, s'_i)$ denote a scaling tensor which stores the weight gradient in the $i$-th GPU, where $g'_i$ is a FP8 tensor and $s'_i$ is the corresponding scaling factor. The actual weight gradient is ${g'_i} / {s'_i}$.
Prior to the all-reduce operation over gradient tensors,  we first gather the scaling factors $s'_{i}$ of each gradient tensor on all GPUs and calculate the global minimum scaling factor $s'_{g}$ as:
\begin{equation}
s'_{g} = \textrm{min}\left(s'_{1},~ s'_{2},~ \ldots,~ s'_{N}\right),
\end{equation}
where the global minimum scaling factor $s'_{g}$ is shared across GPUs. We use this shared scaling factor $s'_g$ to unify the rescaling of the gradient tensors across GPUs. In this way, all gradient tensors associated with the same weight use the same shared scaling factor to quantize the tensors into FP8 format on all GPUs:
\begin{equation}
g''_{i} = \textrm{FP8}\left(s'_{g} \cdot \left(g'_{i} / s'_{i}\right)\right).
\label{eq.global}
\end{equation}
This approach reduces communication overhead by transmitting only a single scalar $s'_{g}$, making the additional synchronization step highly efficient. As the input tensors share the same scaling factor, it eliminates the need of considering all-reduce the scaling factors in parallel and allows for standard NCCL all-reduce operation to be performed. The final collected gradient is obtained as follows: 
\begin{equation}
g = g''_{1} + g''_{2} + \cdots + g''_{N}, \qquad s = N \cdot s'_{g},
\end{equation}
where $g$ is the final aggregated gradient and $s$ is the corresponding scaling factor. Rescaling the scaling factor for the summed gradient $g$ is equivalent to dividing $g$ by $N$ in theory. 
By implementing the aforementioned dual strategies of distributed and automated scaling, we can successfully realize FP8 low-bit gradient communication while preserving model accuracy. Furthermore, this approach  entails storing gradients in FP8 and conducting communication in FP8 as well, thereby yielding reductions in GPU memory usage and communication bandwidth consumption.

\subsection{FP8 Optimizer}

In the training of LLMs, Adam and its variants \citep{adam, adamw} are the most frequently-used optimization methods, that maintain copies of model weights, gradients, first-order and second-order gradient moments for model updates. %
Mixed-precision training \citep{amp} with Adam optimizer typically stores master weights, gradients and gradient moments in 32-bit float format for numerical stability \citep{megatron-lm,zero-deepspeed, opt, bloom}. Consequently, the Adam optimizer consumes 16 bytes of memory per parameter during training:
\begin{equation}
\underbrace{4}_{\text{master weights}} 
+ \underbrace{4}_{\text{gradients}}
+ ~~\underbrace{~~4 ~~+ ~~ 4~~}_{\text{Adam states}}
~~ = ~~ 16\ \text{bytes}.
\label{eq:16bitamp}
\end{equation}
When model size is large, the memory consumption of the variables in Adam will become a bottleneck. %
Previous work \citep{gopher, glm-130b, swinv2} has revealed that reducing precision of the variables in optimizer to 16-bit leads to accuracy degradation when training billion-scale models\footnote{BF16 lacks the precision needed for accuracy, while FP16 has a restricted dynamic range. Given these challenges, prevalent mixed-precision training methodologies rely on utilizing FP32 full-precision for master weights, gradients, and gradient moments.}.
This prompts an evaluation of which variables in the optimizer should be allocated high precision and which can be accommodated with low-precision.

To clarify, we decouple the influence of data precision on the variables in the optimizer and investigate which one can be assigned lower precision, \emph{i.e.}, \emph{precision decoupling}. We find a guiding  principle:  
the gradient statistics can use lower precision, while the master weights necessitate high precision. 
More concretely, the first-order gradient moment can tolerate a high quantization error and can be assigned with low-precision FP8, while the second-order moment requires a higher precision, as analyzed in Sec. \ref{sec.ablation}. This stems from the fact that, during model updates in Adam, the direction of the gradient holds greater significance than its magnitude. 
FP8 with tensor scaling can effectively preserve the distribution of the first-order moment as the high-precision tensor, though it introduces precision degradation to some extend. Calculating the square of gradients for the second-order gradient moment might lead to data underflow due to the typically small gradient values. Therefore, allocating a 16-bit higher precision is necessary to preserve numerical accuracy.

On the other hand, we find that it is crucial to keep the master weights using high precision. %
The underlying reason is that weight updates can sometimes become extremely small or large during training, higher precision for the master weights helps prevent loss of information when updating weights, ensuring more stable and accurate training. In the implementation, the master weights have two viable options: utilizing either FP32 full-precision or FP16 with tensor scaling. FP16 with tensor scaling offers the advantage of conserving memory without compromising accuracy. Consequently, our default choice is to employ FP16 with tensor scaling for storing master weights in the optimizer. Our FP8 mixed-precision optimizer consumes 6 bytes of memory per parameter during training:
\begin{equation}
\underbrace{2}_{\text{master weights}} 
+ \underbrace{1}_{\text{gradients}}
+ ~~\underbrace{~~1 ~~+ ~~ 2~~}_{\text{Adam states}}
~~ = ~~ 6\ \text{bytes}.
\end{equation}
This new low-bit optimizer reduces memory footprints by 2.6x in comparison to the previous solution, as exemplified in Eq.~(\ref{eq:16bitamp}). Noteworthily, this is the first FP8 optimizer for LLM training. The experiments in Sec.~\ref{sec.mainresults} show that FP8 optimizer can preserve model accuracy at various scales, ranging from 125M to 175B parameters.

\subsection{FP8 Distributed Parallel Training}

Training LLMs like GPT-3 requires distributed learning strategies for parallelizing across GPUs. The frequently-used strategies include data parallelism, tensor parallelism, pipeline parallelism, and sequence parallelism. Each parallelism has its own merits %
and has been used in a complementary fashion in existing systems \citep{megatron-nlg, megatron-lm, opt, bloom, colossal-ai}. %
For FP8 supports of these strategies, neither data parallelism nor pipeline parallelism necessitates any specific modifications, because they do not involve additional FP8 compute and communication when splitting data batches or model layers into segments across devices.

\begin{figure}[!t]
\vspace{-4mm}
\centering
\includegraphics[width=0.98\textwidth]{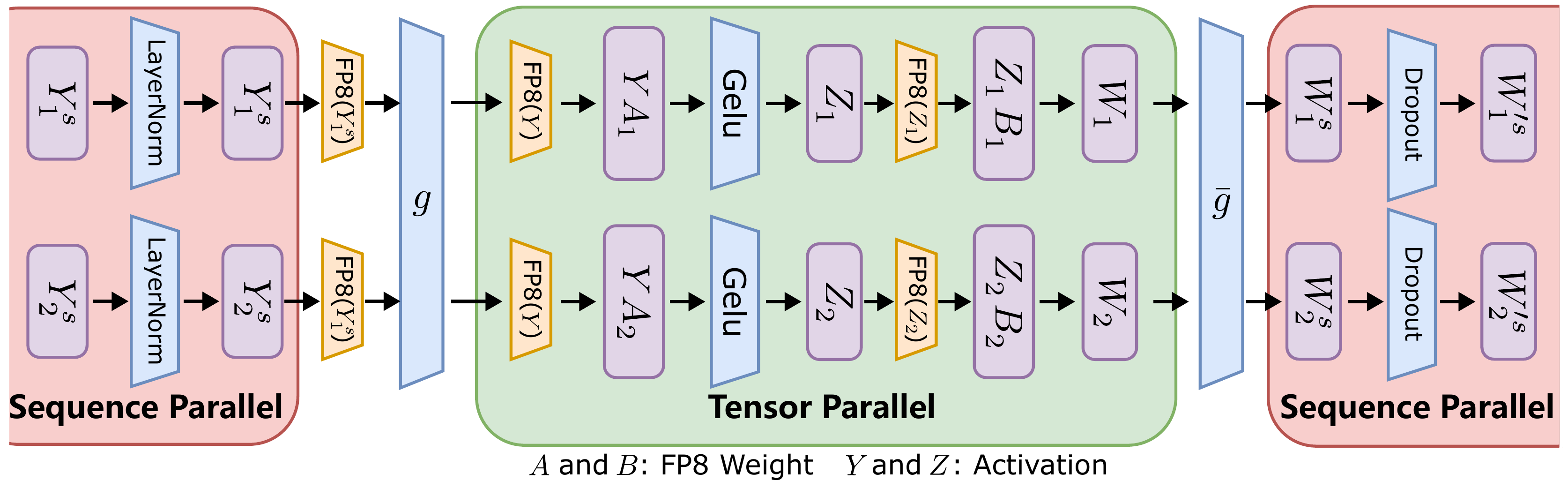}
\caption{Transformer layer with FP8 tensor and sequence parallelism. The FP8 low-bit  operation is highlighted with \textcolor{orange}{orange}. $g$ is all-gather in forward pass and reduce-scatter in backward pass, while $\bar{g}$ is reduce-scatter in forward pass and all-gather
in backward pass. The gather-reduce operation $g$ between sequence parallel and tensor parallel is executed utilizing  FP8 low-precision activation, thus saving half communication costs.}
\label{fig:distribute}

\end{figure}

Tensor parallelism partitions individual layers of a model across multiple devices, such that the shards of weight, gradient and activation tensors are placed on separate GPUs, instead of a single one. To equip tensor parallelism with FP8, we convert the sharded weight and activation tensors to FP8 format for linear layer computation, enabling the forward compute and backward gradient collective communication all using FP8. %
On the other hand, sequence parallelism splits input sequences into multiple chunks and the sub-sequences are fed to different devices to save activation memory. As shown in Fig. \ref{fig:distribute}, sequence and tensor parallelism are performed in parallel to different parts of a Transformer model to make the best use of the available memory and improve training efficiency. There is a converter $g$ between sequence and tensor parallel regions to all-gather sequence partitions in the forward pass (or reduce-scatter tensor segments in the backward pass). We add an FP8 datatype conversion prior to $g$, such that the all-gather (or reduce-scatter) operation uses FP8 low-bit activation to save communication cost across GPUs.

\begin{figure}[!t]
\centering
\includegraphics[width=0.92\textwidth]{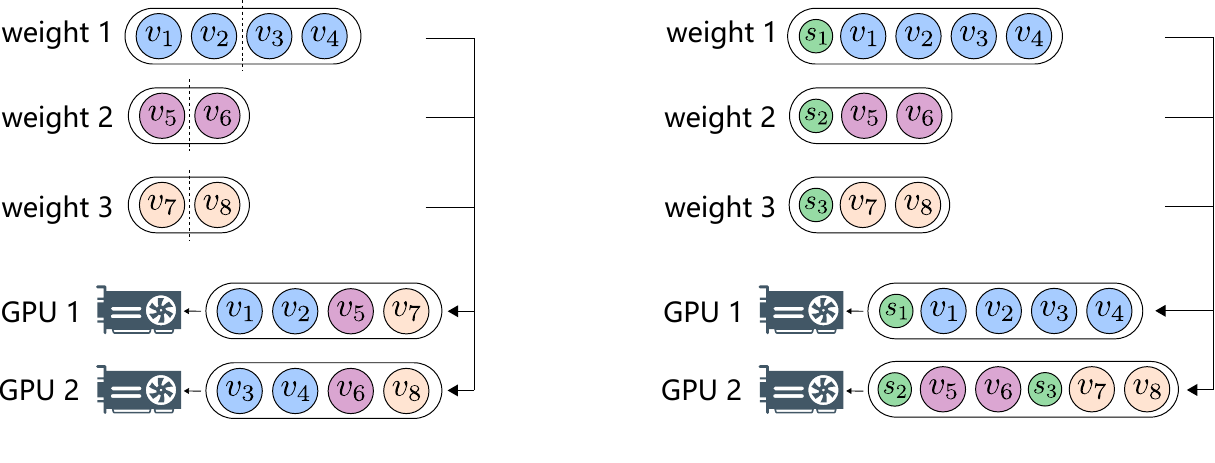}
\vspace{-3mm}
\caption{ZeRO tensor partitioning with and without scaling factors. Left: the original high-precision ZeRO method, which splits a single tensor into multiple partitions and distributes them to different devices. Right: the proposed FP8 ZeRO, which distributes each tensor in its entirety across devices while taking tensor scaling into account.}
\label{fig:scalingZeRO}
\vspace{-5mm}
\end{figure}

In addition, Zero Redundancy Optimizer (ZeRO) \citep{zero-deepspeed} is another frequently-used distributed learning technique in large model training. The core idea of ZeRO is to shade model states over devices, such that each device only hold a fraction of data (\emph{e.g.}, master weights, gradients, and optimizer states) required for a training step. To reduce memory consumption, ZeRO method generally splits a single tensor into multiple partitions and distributes them to different devices. Directly applying FP8 to ZeRO is infeasible, because it is difficult to handle the scaling factors associated with the FP8 partitions. The per-tensor scaling factors should be distributed along with FP8 partitions. To address this issue, we implement a new FP8 distribution scheme that distributes each tensor as a whole across devices, rather than partitioning it into multiple sub-tensors as in ZeRO. The distribution of FP8 tensors is processed in a greedy manner, as outlined in Alg. \ref{alg:greedy}.  
Specifically, our method first sorts the tensors of model states according to their sizes, and then distributes the tensors to different GPUs based upon the remaining memory size of each GPU. The distribution follows the principle that the GPUs with larger remaining memory get a higher priority in receiving new distributed tensors. In this way, the tensor scaling factors can be distributed along with the tensors smoothly, while reducing communication and compute complexity. Figure \ref{fig:scalingZeRO} presents a visual illustration of the difference in ZeRO tensor partitioning between scenarios with and without scaling factors.

\begin{algorithm}[!b]
    \small
    \centering
    \caption{\small{Greedy Distribution Algorithm for ZeRO}}
    \label{alg:greedy}
    \begin{algorithmic}[1]
        \Require FP8 tensors with their corresponding scaling factors: $T=\{(s_1, t_1),(s_2, t_2),\dots,(s_n, t_n)\}$, where $s$ denotes scaling factors while $t$ represents 8-bit tensors. The size of each tensor: $C=\{c_1,c_2,\dots,c_n\}$.
        \Ensure Partitions representing scaling tensors assigned to each GPU.
        \State  Sort $T$ in descending order of their sizes to get $T'=\{(s'_1, t'_1),(s'_2, t'_2),\dots,(s'_n, t'_n)\}$ and $C'=\{c'_1,c'_2,\dots,c'_n\}$, where $c'_1\geqslant c'_2\geqslant\dots\geqslant c'_n$.
        \State Initialize memory usage $u_j=0$ and partition $p_j=\emptyset$ for each GPU $G_j$.
        \For{$i=1$ to $n$}
            \State $j \gets \arg\min_{j} u_j$ \Comment{Find the GPU $j \in [1,m]$ with the least memory usage.}
            \State $p_j\gets p_j\cup \{(s'_i,t'_i)\}$ \Comment{Assign $(s'_i,t'_i)$ to $G_j$.}
            \State $u_j \gets u_j+ c'_i$ \Comment{Update the memory usage of $G_j$.}
        \EndFor
        \State \Return Partitions $P=\{p_1,p_2,\dots,p_m\}$
    \end{algorithmic}
\end{algorithm}

\section{Experiment}

In this section, we assess the effectiveness of the proposed FP8 mixed-precision training approach on GPT-style LLMs, including a wide range of model scales, from 125 million to 175 billion parameters. For performance ablation, we compare GPT models trained with FP8 against those trained with half-precision BF16 and full-precision FP32. For generality evaluation, we conduct experiments encompassing both FP8 low-bit pre-training and fine-tuning, considering instruction tuning and human preference alignment.

\subsection{Experimental Setup}

\subsubsection{Training Dataset}

Our pre-training data is constructed using open-sourced language collections from several sources, including CommonCrawl\footnote{https://commoncrawl.org}, The Pile \citep{pile}, C4 \citep{c4}, OpenWebText \citep{gpt2, Gokaslan2019OpenWebTextCorpus}, CC-NEWS \citep{liu2019roberta}, CC-Stories \citep{trinh2018simple}, Redpajama \citep{redpajama}, and Wikipedia\footnote{https://wikipedia.org}. 
We apply fuzzy deduplication \citep{lee_deduplicating_2022} across CommonCrawl snapshots to enhance data quality. %
Tab. \ref{tab.data} in Appendix \ref{appendix.data} provides details of our pre-training data, including information such as the number of tokens from each source and associated sampling weights. 
For a more comprehensive understanding of the data and its cleaning pipeline, readers are encouraged to refer to Appendix \ref{appendix.data}. 

Moreover, for instruction tuning, we follow the same settings as Vicuna-v1.1\citep{vicuna}, which uses a publicly user-shared instruction following data \citep{sharegpt}. %
For reinforcement learning with human feedback, the training data we used is a combination of the Anthropic's Helpful and Harmless dataset~\citep{HH} and Open-Assistant dataset~\citep{oasst}. The training framework and associated configurations align with the publicly available AlpacaFarm~\citep{alpacafarm}.

\subsubsection{Model Configuration}

\begin{table}[t!]
\center
\begin{tabular}{cccccccccc}
\toprule
params & dimension & $n$ heads & $n$ layers & TP & PP & SP & learning rate & batch size & $n$ tokens \\
\midrule\
125M  & 768 & 12 & 12 & 1 & 1 & \checkmark & $6.0e^{-4}$ & 1M & 100B \\
7B  & 4096 & 32 & 32 & 1 & 1 & \checkmark & $3.0e^{-4}$ & 4M & 100B \\
13B & 5120 & 40 & 40 & 2 & 1 & \checkmark & $3.0e^{-4}$ & 4M & 100B \\
175B & 12288 & 96 & 96 & 8 & 4 & \checkmark &  $3.0e^{-5}$ & 1M & 40B \\
\bottomrule
\end{tabular}
\caption{
{Model sizes, architectures, and training hyper-parameters. TP, PP, and SP indicate tensor, pipeline, and sequence parallelism, respectively. %
To mitigate carbon emissions and save cost, we restrict the training of the 175B model to a dataset comprising  only 40B tokens, which has proven to be sufficient for evaluating system performance.}
\label{tab.arch}
\vspace{-8mm}
}

\end{table}

The model architecture we used is a decoder-only Transformer \citep{gpt3}, which has been widely-used in recent generative LLMs like PaLM \citep{palm}, OPT \citep{opt}, and LLaMA \citep{llama}. In addition to the base architecture, we integrate several modifications proposed recently to improve model efficiency and effectiveness. %
1) \emph{Rotary Positional Embedding}: Drawing inspiration from recent successful experiments \citep{gpt-neox, llama}, we incorporate rotary positional embeddings (RoPE) \citep{rope} into our approach. This addition enables us to capture both absolute and relative positions information, enhancing performance especially when extrapolating to larger context windows.
2) \emph{Flash Attention}: The standard attention implementation is bottlenecked by memory access \citep{ivanov2021data}. Flash Attention \citep{flashatt} proposed an IO-aware exact attention algorithm which uses tiling to reduce the amount of HBM accesses, achieving substantial acceleration.

We train the models using the proposed FP8 optimizer, which is built upon Adam \citep{adam} with decoupled weight decay \citep{adamw}, following the common practise with the decay rates $\beta_1$ = 0.9, $\beta_2$ = 0.95, and weight decay = 0.1. The learning rate schedule is cosine-like, and the final learning rate is 10\% of the maximal learning rate. We train the models for 100B tokens in total with a batch size of 4M tokens, and the input sequence length is set to 2048. The model warm-up is conducted for 1,000 iterations. Tab. \ref{tab.arch} presents the details of model configurations and the corresponding training settings. The training is conducted on Azure NDv5 H100 GPU platform \citep{hpc}.

\begin{figure}[!t]
\begin{minipage}[t]{1.0\textwidth}
\centering
\subfloat[][GPT-7B]{
\includegraphics[height=3.4cm]{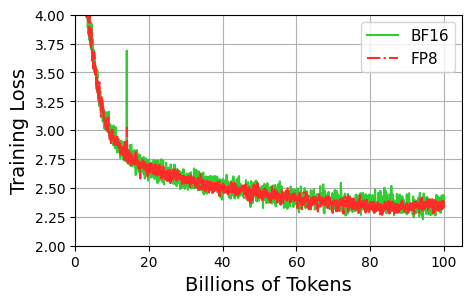}
}
\centering
\subfloat[][GPT-13B]{
\includegraphics[height=3.4cm]{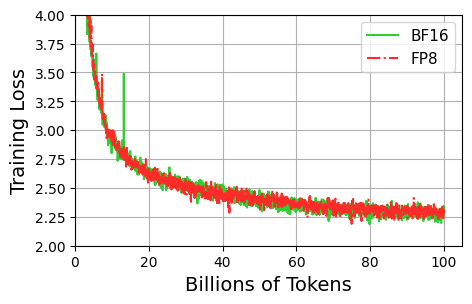}
}
\centering
\subfloat[][GPT-175B]{
\includegraphics[height=3.4cm]{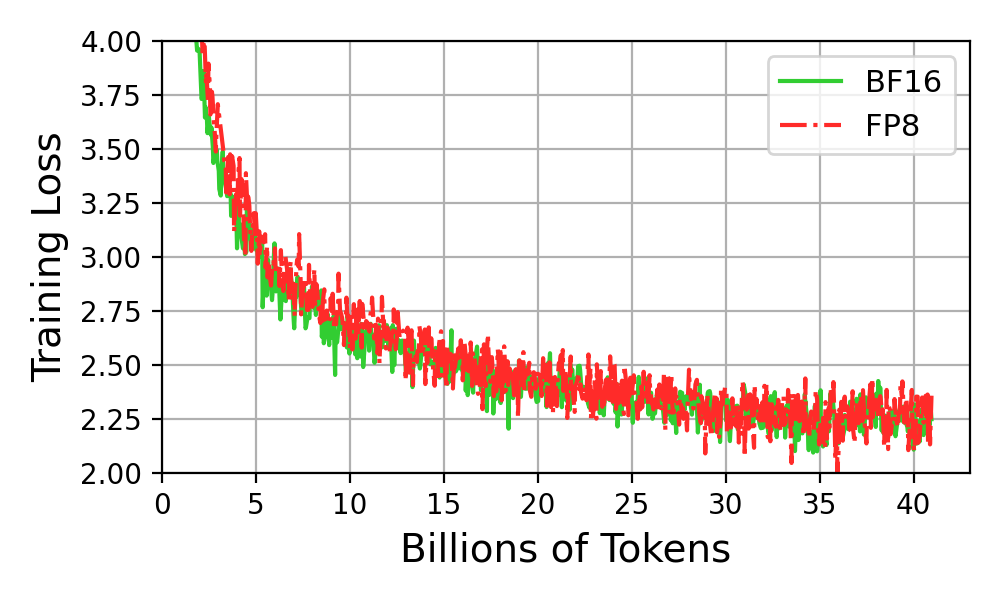}
}
\end{minipage}
\caption{A comparison between FP8 and BF16: Analyzing the training loss of GPT models with the parameters ranging from 7 billion to 175 billion.}
\label{fig:gptloss}
\end{figure}

\begin{table}
\vspace{-2mm}
\centering
\scalebox{0.95}{
\begin{tabular}{@{}lcccccccccc@{}}
\toprule
& ~~~\textbf{HS} & \textbf{Lambada} & \textbf{BoolQ} & \textbf{PIQA} & \textbf{COPA} & \textbf{Winogrande} & \textbf{Arc-C} & \textbf{Arc-E} & \textbf{ObQA} & \bf Avg \\
\midrule
\multicolumn{8}{l}{\textit{GPT-7B model zero-shot performance}} & \\
\bf{BF16} & 61.3 & 61.4 & 61.2 & 75.0 & 79.0 & 58.5 & 32.9 & 59.7 & 36.4 &  58.4 \\
\bf{FP8} & 60.0 & 61.8 & 62.0 & 74.2 & 78.0 & 59.8 & 32.9 & 58.7 & 34.6 &  58.0 \\
\midrule
\multicolumn{8}{l}{\textit{GPT-13B model zero-shot performance}} & \\
\bf{BF16} & 64.8 & 64.9 & 63.4 & 75.9 & 82.0 & 61.0 & 35.2 & 61.5 & 40.6 &  61.0 \\
\bf{FP8} & 64.1 & 63.4 & 63.9 & 76.2 & 81.0 & 61.6 & 34.9 & 61.3 & 36.8 &  60.4 \\
\bottomrule
\end{tabular}
}
\caption{Zero-shot performance on downstream tasks. The models are trained with either the standard BF16 mixed-precision scheme \citep{megatron-lm} or the proposed FP8 low-precision scheme. 
}
\label{tab.downstreamtask}
\vspace{-1.5em}
\end{table}

\subsection{Main Results}
\label{sec.mainresults}

\subsubsection{Model Performance}
We first compare the performance of models trained using FP8 mixed-precision with those trained using BF16. In Fig.~ \ref{fig:gptloss}, the pre-training loss over tokens is displayed for GPT models of 7B, 13B, and 175B parameters. The training configurations and hyper-parameters remain consistent across models trained with FP8 and BF16. %
The only difference lies in the mixed-precision schemes utilized. As shown in Fig. \ref{fig:gptloss}, the loss curves almost overlap with each other. %
The results unequivocally demonstrate that the proposed FP8 mixed-precision scheme can achieve equivalent performance to the prevalent higher-precision BF16 scheme \citep{megatron-lm, gopher, chinchilla} across a diverse array of model scales. 
Also, we evaluate the pre-trained models on a wide range of downstream tasks, including HellaSwag (HS) \citep{hellaswag}, Lambada \citep{lambada}
BoolQ \citep{boolq}, PIQA \citep{piqa}, COPA \citep{copa}, Winogrande \citep{winogrande}, Arc \citep{arc}, and OpenbookQA (ObQA) \citep{openbookqa}. As reported in Tab. \ref{tab.downstreamtask}, the FP8 pre-trained models exhibit  comparable zero-shot performance in comparison to their BF16 counterparts. This result provides further validation that models pre-trained with FP8 low-precision maintain both accuracy and intrinsic in-context learning capabilities at a level comparable to their high-precision counterparts.

\begin{figure}[!t]
\begin{minipage}{\textwidth}

\begin{minipage}[b]{0.4\textwidth}
\centering
\includegraphics[height=3.5cm]{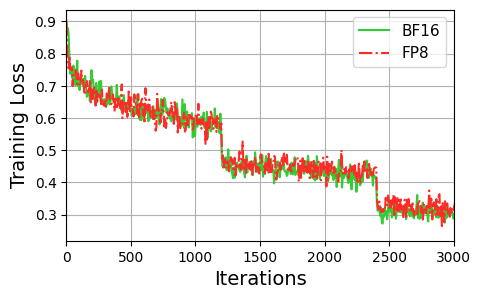}
\vspace{-0.5em}
\captionof{figure}{SFT training loss.} %
\label{fig.sft}
\end{minipage}
\hfill
\begin{minipage}[b]{0.59\textwidth}
\centering
\vspace{-5em}
\scalebox{0.8}{
\begin{tabular}{@{}l|cc|cc@{}}
\toprule
\multirow{2}{*}{Mixed-precision} & \multicolumn{2}{c|}{System Performance} & \multicolumn{2}{c}{Model Performance} \\ \cmidrule(l){2-5} 
                                 & GPU Mem. (GB)        & Throughput       & AlpacaEval     & MT-Bench            \\ \midrule
\bf{BF16}                             & 51.1        & 103                 & 66.15           & 5.75    \\
\bf{FP8}                              & 44.0({\scriptsize{\textcolor{myyellowgreen}{-14\%}}})        & 131({\scriptsize{\textcolor{myyellowgreen}{+27\%}}})                  & 67.20           & 5.70    \\ \bottomrule
\end{tabular}
}
\vspace{-0em}
\captionof{table}{A comparison between FP8 and BF16 for SFT. For system performance, we report results of GPU memory usage and training throughput. For model performance, we present the win-rate against Davinci-003 on AlpacaEval and GPT-4 judged scores on MT-Bench.}
\label{tab.sft}
\end{minipage}

\hspace{1em}
\end{minipage}
\vspace{-1em}
\end{figure}

Furthermore, we leverage the proposed FP8 mixed-precision approach for fine-tuning LLMs in instruction following. For a fair comparison, we follow the same instruction tuning settings as Vicuna-v1.1 \citep{vicuna}, which adopts the open-sourced LLaMA-7B \citep{llama} as the base model for fine-tuning. Fig. \ref{fig.sft} presents the fine-tuning loss, where the curves corresponding to BF16 and FP8 display a notable degree of overlap. Meanwhile, the win-rate of our FP8 fine-tuned models against Davinci-003 \citep{davinci-003} is also comparable to that of Vicuna-v1.1, which is fine-tuned using BF16 half-precision, as reported in Tab. \ref{tab.sft}. This indicates that our FP8 low-bit training scheme is versatile, as it is applicable not only to pre-training phase but also to downstream fine-tuning tasks.

\begin{figure}[!t]
\begin{minipage}[!t]{\textwidth}
\begin{minipage}[b]{0.4\textwidth}
\centering
\includegraphics[height=3.5cm]{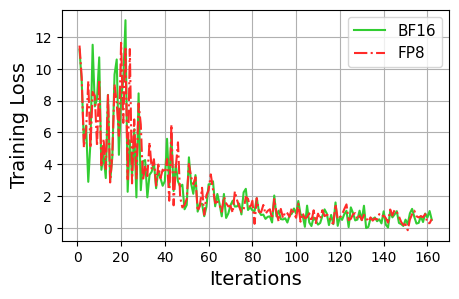}
\vspace{-0.5em}
\captionof{figure}{RLHF training loss.}
\label{fig.rlhf}
\end{minipage}
\vspace{-3mm}
\begin{minipage}[b]{0.58\textwidth}
\centering
\scalebox{0.8}{
\begin{tabular}{@{}l|cc|cc@{}}
\toprule
\multirow{2}{*}{Mixed-precision} & \multicolumn{2}{c|}{Memory Usage (MB)} & \multicolumn{2}{c}{Model Performance} \\ \cmidrule(l){2-5} 
                                 & Weights        & Optimizer States       & AlpacaEval     & MT-Bench            \\ \midrule
\bf{BF16}                             & 15,082        & 15,116                 & 72.05           & 6.16    \\
\bf{FP8}                              & 10,292({\scriptsize{\textcolor{myyellowgreen}{-32\%}}})        & 5,669({\scriptsize{\textcolor{myyellowgreen}{-62\%}}})                  & 72.42           & 6.04    \\ \bottomrule
\end{tabular}
}
\captionof{table}{A comparison of FP8 and BF16 RLHF alignment. Memory usage is assessed with a focus on weights and optimizer states, while model performance is evaluated on AlpacaEval considering win-rate against Davinci-003, and MT-Bench using GPT-4 judged scores.}
\label{tab.rlhf}
\end{minipage}
\end{minipage}
\end{figure}

In addition, we further apply the proposed FP8 mixed-precision scheme to reinforcement learning from human feedback (RLHF), a more complex process to align LLMs with user preferences. %
Following the same training setting as AlpacaFarm \citep{alpacafarm}, a recent RL framework for LLM alignment, we optimize policy models with PPO algorithm \citep{ppo}. The solely difference lies in the choice of mixed-precision training schemes, \emph{i.e.}, BF16 \emph{v.s.} FP8.  
From the results reported in Fig. \ref{fig.rlhf} and Tab. \ref{tab.rlhf}, we observe a notable reduction in memory utilization, for instance, a 32\% memory reduction concerning model weights and a 62\% reduction concerning optimizer states. Consequently, it can be inferred that FP8 is capable of replicating the BF16 mixed-precision for RLHF training. This  underscores the broader applicability and versatility of our FP8 low-bit training solution. %

\vspace{-2mm}

\subsubsection{System Performance}

In this section, we evaluate system-level performance of FP8 mixed-precision, considering communication efficiency, memory utilization, and the overall speed, with an emphasis on cost savings. Our method employs 8-bit gradients for all-reduce collective communication among GPUs. Theoretically, this results in a 75\% reduction in communication costs when compared to the mainstream 32-bit scheme (Despite BF16 mixed-precision computing gradients using 16-bit precision, it still employs 32-bit precision for all-reduce communication \citep{megatron-lm}). 
Due to the impact of system transmission loss, the observed practical reduction during GPT model training falls within the range of 63\% to 65\%, as indicated in Table \ref{tab.sys}. Furthermore, it is worth noting that the recent Nvidia Transformer Engine (TE) \citep{te} still relies on full-precision FP32 for collective communication, %
resulting in the same level of reduction for our FP8 solution.

When training GPT models with identical batch sizes, FP8 mixed-precision can lead to a reduction in memory footprint ranging from 28\% to 39\% when compared to BF16, as reported in Tab. \ref{tab.sys}. These reductions in memory consumption are attributed to the FP8 gradient and FP8 optimizer techniques we have introduced. Moreover, compared with TE \citep{te}, our solution is also very competitive, obtaining 36.1\%, 36.0\%, and 42.1\% additional memory reductions for different model sizes, \emph{i.e.}, GPT-7B, 13B, and 175B. Although TE employs FP8 for compute, it still uses high-precision optimizer and gradients, which consumes much more memory than our solution. In addition, the saved memory in our method can be used to train larger batch size or longer sequence. For example, when employing 32 H100 GPUs with a memory capacity of 80GB, our approach enables the training of models with a context of 4,096 tokens, accommodating up to 175 billion parameters. In contrast, TE can only accommodate models with a context of 2,048 tokens. This showcases the potential of integrating our FP8 mixed-precision training into existing LLMs, empowering them to train longer sequences with the same GPU resources.%

Moreover, our FP8 mixed-precision scheme shows a superior training throughput compared to the prevalent BF16 scheme, achieving a notable speed-up of 75\% when applied to GPT-175B model. The model FLOPS utilization (MFU) of FP8 mixed-precision training is 34.2\% on H100 GPUs, being 37.3\% superior to TE. These findings provide substantial evidence that our FP8 scheme effectively conserves memory, reduces communication costs during the training of large models, and ultimately enhances system utilization efficiency on the latest H100 GPU platform.

\begin{table}[!t]
\vspace{-1mm}
\centering
\scalebox{1}{
\resizebox{\textwidth}{!}{%
\begin{tabular}{@{}lccccccccccc@{}}
\toprule
\multirow{2}{*}{Model} & \multirow{2}{*}{TP} & \multirow{2}{*}{PP} & \multirow{2}{*}{DP} & Micro & Mixed & GPU & Throughput & \multirow{2}{*}{TFLOPS} & MFU & \multicolumn{2}{c}{Weight-related Comm.} \\
~ & ~ & ~ & ~ & BS & Precision & Mem. (GB) & (\#samples/s) & ~ & (\%) & Rate (\%) & Volume (GB) \\
\midrule

\multirow{4}{*}{GPT-7B} & \multirow{4}{*}{1} & \multirow{4}{*}{1} & \multirow{4}{*}{32} & 2 & BF16 & 69.6 & 159.2 & 445 & 45.0 & 10.1 & 37.2 \\
& & & & 2 & FP8 (TE) & 77.3 & 224.5 & 627 & 31.7 & 9.7 & 37.2 \\
& & & & 2 & FP8 (Ours) & 49.4 ({\scriptsize{\textcolor{myyellowgreen}{-29\%}}}) & 219.8 ({\scriptsize{\textcolor{myyellowgreen}{+38\%}}}) & 615 & 31.1 & 7.9 &  13.9 ({\scriptsize{\textcolor{myyellowgreen}{-63\%}}}) \\
~ & ~ & ~ & ~ & 4 & FP8 (Ours)  & 69.3 & 230.5 ({\scriptsize{\textcolor{myyellowgreen}{+45\%}}}) & 645 & 32.6 & 10.4 &  13.9 ({\scriptsize{\textcolor{myyellowgreen}{-63\%}}}) \\
\midrule

\multirow{4}{*}{GPT-13B} & \multirow{4}{*}{2} & \multirow{4}{*}{1} & \multirow{4}{*}{16} & 2 & BF16  & 68.2 & 79.3 & 420 & 42.5 & 11.1 &  34.3 \\
& & & & 2 & FP8 (TE)  & 76.4 & 111.7 & 592 & 29.9 & 7.1 &  34.3 \\
& & & & 2 & FP8 (Ours)   & 48.9 ({\scriptsize{\textcolor{myyellowgreen}{-28\%}}}) & 109.5 ({\scriptsize{\textcolor{myyellowgreen}{+38\%}}})  & 575 & 29.1 & 3.9 &  12.4  ({\scriptsize{\textcolor{myyellowgreen}{-64\%}}}) \\
& & & & 4 & FP8 (Ours)   & 67.8 & 121.5 ({\scriptsize{\textcolor{myyellowgreen}{+53\%}}}) & 644 & 32.5 & 9.3 &  12.4 ({\scriptsize{\textcolor{myyellowgreen}{-64\%}}}) \\

\midrule

\multirow{4}{*}{GPT-175B} & \multirow{4}{*}{8} & \multirow{4}{*}{4} & \multirow{4}{*}{4} & 1 & {BF16} & 66.1 & 22.4 & 386 & 39.0 & 8.8 &  23.4 \\
&&&& 1 & {FP8 (TE)}  & 69.6 & 28.7 & 493 & 24.9 & 3.9 &  23.4 \\
&&&& 1 & {FP8 (Ours)}   & 40.3 ({\scriptsize{\textcolor{myyellowgreen}{-39\%}}}) & 27.1 ({\scriptsize{\textcolor{myyellowgreen}{+21\%}}}) & 473 & 23.9 & 2.5 &  8.2 ({\scriptsize{\textcolor{myyellowgreen}{-65\%}}}) \\
&&&& 4 & {FP8 (Ours)}   & 57.7 & 39.3 ({\scriptsize{\textcolor{myyellowgreen}{+75\%}}}) & 677 & 34.2 & 10.9 &  8.2  ({\scriptsize{\textcolor{myyellowgreen}{-65\%}}}) \\

\bottomrule
\end{tabular}%
}
}
\caption{System-level performance on Nvidia H100 GPUs 80GB. Here, TP, PP, and DP represent tensor, pipeline, and data parallelism respectively. BS indicates batch size, while MFU denotes model FLOPs utilization. Weight-related communication contains the all-gather operator on weights and the reduce-scatter operator on weight gradients.}
\label{tab.sys}
\end{table}

\begin{figure*}[t]
\vspace{-1em}
    \begin{minipage}[b]{1.0\textwidth}
    \subfloat[SNR (Signal to Noise Ratio)]{
        \includegraphics[height=2.7cm, trim=0 0 0 0]{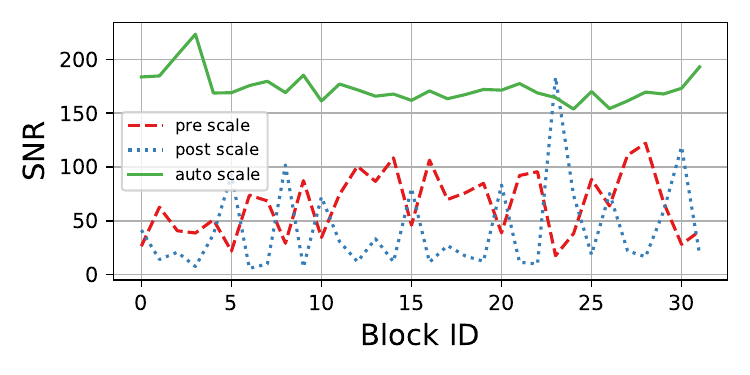}
    }
    \subfloat[Underflow rate]{
        \includegraphics[height=2.7cm, trim=0 0 0 0]{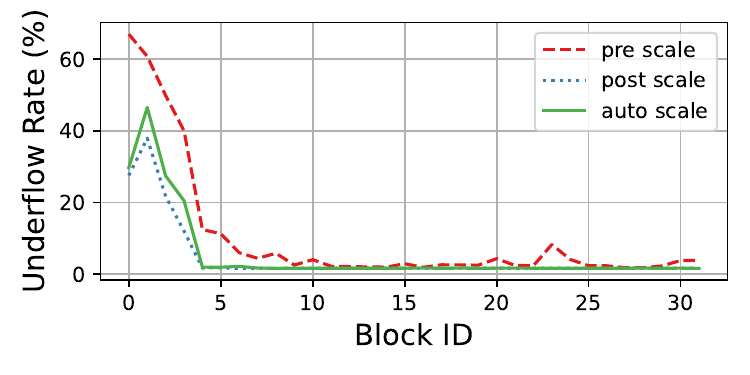}
    }
    \subfloat[Overflow rate]{
        \includegraphics[height=2.7cm, trim=0 0 0 0]{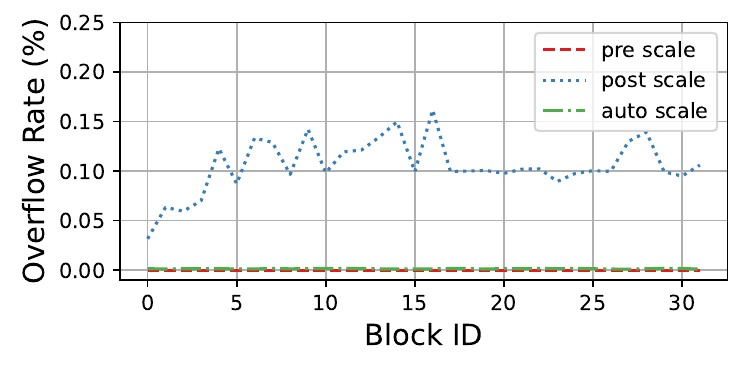}
    }

    \end{minipage}
    \caption {Comparing different strategies, \emph{i.e.}, pre-scaling, post-scaling, and auto-scaling, for FP8 gradient all-reduce.  
    We investigate SNR, underflow rate, and overflow rate across different Transformer blocks. The experiment is conducted using a GPT-7B model with a data parallelism factor of 128.}
    \vspace{-4mm}
    \label{fig:ablate_comm}
\end{figure*}

\subsection{Ablation Study}
\label{sec.ablation}

\begin{figure}%
\begin{minipage}[!t]{\textwidth}
\begin{minipage}[b]{0.47\textwidth}
\centering
\scalebox{0.8}{
\begin{tabular}{cccccccccc}
\toprule
\makecell[c]{Low-bit\\ Settings}  &  \makecell[c]{  Compute \\ (GEMM)}
 & Comm. &  \makecell[c]{Master \\ Weight} & \makecell[c]{Optimizer \\ States}  \\
\midrule
FP32 \textcolor{cyan}{\#0} & FP32 & FP32 & FP32 & FP32+FP32 \\
BF16 \textcolor{blue}{\#1} & \underline{BF16} & FP32  &  FP32 & FP32+FP32  \\
FP8 \textcolor{red}{\#2a} & FP8  &  FP8 & \underline{FP16} & 
\underline{FP8}+\underline{FP16} \\
FP8 \textcolor{pink}{\#2b} & FP8  &  FP8 & \underline{BF16} & 
FP8+FP16 \\
FP8 \textcolor{green}{\#3} & FP8  &  FP8 & \underline{FP8} & FP8+FP16 \\
FP8 \textcolor{orange}{\#4} & FP8  &  FP8 & FP16 & FP8+\underline{FP8} \\
\bottomrule
\end{tabular}
}
\captionof{table}{Precision decoupling for the variables within the optimizer. Here, our focus is on ablating the master weight and optimizer states, as these components are precision sensitive. The optimizer states include both first-order and second-order gradient moments. Note that the FP16 master weight uses tensor scaling.}
\label{tab.opt}
\end{minipage}
\hfill
\begin{minipage}[b]{0.45\textwidth}
\centering
\includegraphics[width=1.0\textwidth]{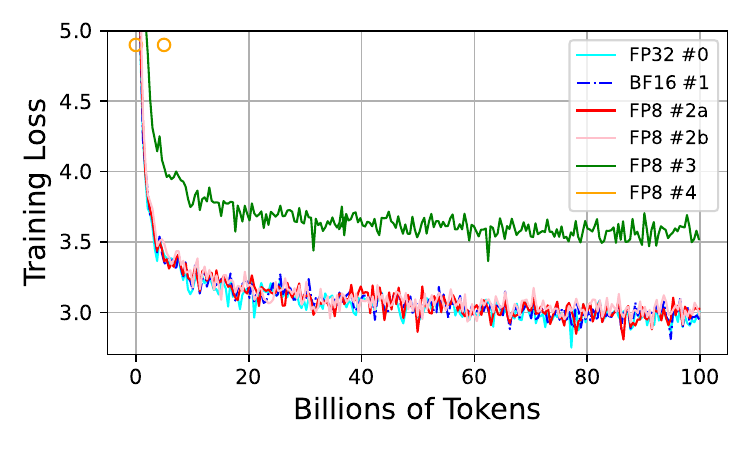}
\vspace{-2mm}
\captionof{figure}{Training losses of GPT-125M models with the settings presented in Tab. \ref{tab.opt}. The loss curve for FP8 \textcolor{orange}{\#4} has diverged.}%
\label{fig.opt}
\end{minipage}
\end{minipage}
\vspace{-4mm}
\end{figure}

We ablate various design choices of FP8 mixed-precision training strategy for LLMs and report the performance in Tab. \ref{tab.opt} -- \ref{tab:zero} and Fig. \ref{fig:ablate_comm} -- \ref{fig.opt}. %
The ablation experiments are conducted on GPT models, whose architectures and training settings are elaborated in Tab. \ref{tab.arch}. Importantly, our ablation study yields several guidelines for the effective utilization of 8-bit datatype in LLM training, which can facilitate future research on low-bit model training.

\textit{Communication}. We first analyze the limitations of the conventional pre-scaling and post-scaling methods when aggregating low-bit gradients during the all-reduce communication process. As shown in Fig. \ref{fig:ablate_comm}, we conduct a statistical analysis on SNR, underflow rate, and overflow rate of weight gradients across different Transformer blocks. It is observed that the pre-scaling method has relative larger underflow rate when quantifying gradients from 32-bit to 8-bit, while the post-scaling method has higher overflow rate. In contrast, the proposed auto-scaling technique can diminish both the underflow ratio and the overflow ratio, while getting much better SNR, as shown in Fig. \ref{fig:ablate_comm} (a). This demonstrates the effectiveness of auto-scaling method in reducing quantization errors when utilizing 8-bit datatype for gradient all-reduce.

\begin{wraptable}{r}{0.5\textwidth}
\centering
\begin{minipage}[t]{0.45\textwidth}
\centering
\scalebox{0.6}{
\begin{tabular}{@{}lccccccc@{}}
\toprule
\multirow{2}{*}{Model} & \multirow{2}{*}{TP} & \multirow{2}{*}{PP} & \multirow{2}{*}{DP}  & Micro & Mixed & \multicolumn{2}{c}{Act-related Comm.} \\
&&&&BS & Precision & Rate (\%) & Volume (GB) \\
\midrule
\multirow{2}{*}{GPT-13B} & \multirow{2}{*}{2} & \multirow{2}{*}{1}& \multirow{2}{*}{16} & \multirow{2}{*}{2} & BF16 & 12.9 & 4.7 \\
&&&&& FP8 (Ours) & 5.3 & 3.1\\

\midrule

\multirow{2}{*}{GPT-175B} & \multirow{2}{*}{8} & \multirow{2}{*}{4} & \multirow{2}{*}{4} & \multirow{2}{*}{1} &  BF16 & 14.9 & 5.9 \\
&&&&& FP8 (Ours) & 5.2 & 3.9\\
\bottomrule
\end{tabular}
}
\caption{Activation-related communication volume reduction in sequence and tensor parallelism, including the all-gather operator on activation and the reduce-scatter on activation gradients.}
\label{tab:sp}
\vspace{1em}
\end{minipage}

\begin{minipage}[t]{0.45\textwidth}
\centering
\scalebox{0.65}{
\begin{tabular}{@{}lccccccccc@{}}
\toprule
\multirow{2}{*}{Model} & \multirow{2}{*}{TP} & \multirow{2}{*}{PP} & \multirow{2}{*}{DP} & Micro & Mixed & \multicolumn{4}{c}{GPU Memory} \\
&&& & BS & Precision & Min & Max   \\
\midrule
\multirow{3}{*}{GPT-7B} & \multirow{3}{*}{1} & \multirow{3}{*}{1} & \multirow{3}{*}{32} & \multirow{3}{*}{2} & BF16 & 69.07 & 69.63  \\
&&&&& FP8 (TE) & 76.97 & 77.28   \\
&&&&& FP8 (Ours) & 49.06 & 49.36   \\
\midrule
\multirow{3}{*}{GPT-13B} & \multirow{3}{*}{2} & \multirow{3}{*}{1} & \multirow{3}{*}{16} & \multirow{3}{*}{2} & BF16 & 67.98 & 68.18   \\
&&&&& FP8 (TE) & 73.68 & 76.36 \\
&&&&& FP8 (Ours) & 48.45 & 48.85   \\
\midrule
\multirow{3}{*}{GPT-175B} & \multirow{3}{*}{8} & \multirow{3}{*}{4} & \multirow{3}{*}{4} & \multirow{3}{*}{1} &  BF16 & 65.60 & 66.12  \\
&&&&& FP8 (TE) & 69.04 & 69.57   \\
&&&&& FP8 (Ours) & 38.64 & 40.28  \\
\bottomrule
\end{tabular}
}
\caption{Comparing ZeRO distribution methods in terms of memory load across GPUs. Here ``Min'' and ``Max'' denote the minimum and maximum memory utilization observed across GPUs. 
Our FP8 ZeRO method uses less memory while achieving memory-aware load balancing.}
\label{tab:zero}
\end{minipage}
\vspace{-2em}
\end{wraptable}

\textit{Optimizer}. We further ablate the impact of reduced precision for the variables in the AdamW optimizer. %
We set the BF16 mixed-precision optimizer as the baseline, since it has been widely used in existing LLM training frameworks \citep{amp, megatron-lm, te}. Tab. \ref{tab.opt} presents the settings of reduced precision for the variables, while Fig. \ref{fig.opt} plots the corresponding training losses. We observe that: 1) FP8 master weight induces performance degradation (see the \textcolor{red}{\#2a} \emph{vs.} \textcolor{green}{\#3} lines in Fig. \ref{fig.opt}), while FP16 can maintain accuracy as FP32 (see \textcolor{red}{\#2a} \emph{vs.} \textcolor{cyan}{\#0} and \textcolor{blue}{\#1}) but requiring using tensor scaling. It reveals that the master weight is precision-sensitive. This can be attributed to the master weight's role in updating weights, which tend to exhibit small magnitudes, necessitating  high precision to maintain accuracy.
2) The training loss of BF16 master weight is slightly higher than that of FP16 with a scaling factor because BF16 has fewer mantissa bits, resulting in lower precision (see \textcolor{red}{\#2a} \emph{vs.} \textcolor{pink}{\#2b}).
3) The second-order gradient moment is more precision-sensitive than the first-order one, because the square calculation is easy to cause underflow and leads to accuracy degradation. Utilizing FP8 for the second-order gradient moment can lead to divergent training loss (see the \textcolor{orange}{\#4} dot in Fig. \ref{fig.opt}). 

\textit{Parallelism}. In our FP8 LLM training framework, we introduce FP8 low-bit convertors into sequence parallelism and tensor parallelism to reduce activation communication costs across GPUs. Here we conduct an analysis experiment to count the activation-related communication volume during GPT model training, and report the numbers in Tab. \ref{tab:sp}. It is observed that our FP8 parallel scheme results in a substantial reduction of 34\% in activation-related communication costs compared to the original method utilizing BF16. Furthermore, in ZeRO distributed training, our method distributes each FP8 tensor along with its associated scaling factor as a whole, rather than partitioning the tensor into splits across GPUs. This strategy not only results in more GPU memory savings but also maintains a balanced memory load across GPUs, as demonstrated in Tab. \ref{tab:zero}.

\section{Related Work}
\textbf{Mixed-precision Training.} Efficient training through reduced mixed-precision has been widely used in modern deep learning to save computing costs. While some works have taken bit-reduction to the extreme, \emph{i.e.} 1-bit binary networks \citep{hubara2016binarized, Xnor-net}, they have not been successful in maintaining model accuracy \citep{fp8-dl}. The most practical scheme now is the FP16 half-precision method \citep{amp}, which can maintain accuracy while improving training efficiency. %
The computations during forward pass and back propagation use FP16 while the master weights use FP32. Since FP16 has a narrower
dynamic range, FP16 mixed-precision entails loss scaling \citep{amp} to prevent loss of accuracy. Fortunately, the need for loss scaling can be avoided by using BF16 datatype, because BF16 maintains the same dynamic range as the full-precision FP32. This results in that large model training now prefers to use BF16 mixed-precision scheme, which is more stable during training \citep{megatron-nlg, bloom, glm-130b}. %

FP8 is a natural progression from 16-bit data formats to further reducing computing cost. Early pioneering efforts in FP8 low-bit model training \citep{fp8-wangnaigang, sunxiao-fp8, 8bit-opt} have largely remained at the simulation stage. %
Consequently, there exists a notable gap between the projected capabilities of these approaches and their actual performance on hardware \citep{fp8-dl}. 
With the advent of Nvidia Hopper GPU architecture \citep{h100-whitepaper}, FP8 is emerging as a viable and practical data type for the next-generation low-precision training, as discussed in \citep{fp8-dl}. %
At present, the Nvidia Transformer Engine (TE) \citep{te} serves as the primary framework for FP8 mixed-precision training. However, its support for FP8 usage remains somewhat constrained. TE's current implementation restricts FP8 usage solely to weight computation, retaining the storage of model weights and gradient calculations with 16-bit data types. 
Consequently, the end-to-end speed-up, memory and communication cost savings are limited. In contrast, our work infiltrates FP8 gradient, optimizer, and distributed training into the whole progress of model training, fully unveiling the capabilities of FP8. %

\textbf{Large Language Models.} Recent years have witnessed a substantial evolution in the field of LLMs. Autoregressive language modeling -- predicting the future of a text sequence from its past -- provides a simple yet powerful objective that admits formulation of numerous tasks. While there exist alternative methodologies, such as masked language modeling \citep{bert} and permutation language modeling \citep{xlnet}, the autoregressive method now is more promising because of its strong performance. Following the scaling laws \citep{gpt3} and the refined laws \citep{chinchilla}, various  LLMs are have been proposed, including dense models: GPT-3 \citep{gpt3}, Jurassic-1 \citep{jurassic}, Gopher~\citep{gopher}, Chinchilla \citep{chinchilla},  Bloom~\citep{bloom}, OPT \citep{opt} Megatron-Turing NLG \citep{megatron-nlg}, PaLM \citep{palm}, LaMDA \citep{lamda}, LLaMA \citep{llama}, and sparse models: GLaM \citep{glam}, and Switch transformers \citep{switch-trans}. Each of them has demonstrated remarkably competitive few-shot performance across a wide range of tasks at the time of their respective releases. Nonetheless, these models still encounter challenges, such as overwhelming computational requirements  and the need for acquiring more high-quality training data. In this work, we delve into  the utilization of low-precision techniques to mitigate the training costs,  which is a crucial step for the continued expansion of language models.

Low-precision training has been widely used in LLM training to reduce compute cost. OPT \citep{opt} and GLM \citep{glm-130b} utilize FP16 for forwards and backwards and FP32 for optimizer states and master weights, to reduce the GPU memory usage and improve training efficiency. Bloom \citep{bloom} find that FP16 can cause numerical instabilities and irreversible divergences, especially when training models larger than 100B parameters, because FP16's dynamic range is limited. Consequently, Bloom and other LLMs, such as Gopher \citep{gopher} and Chinchilla \citep{chinchilla}, adopt BF16 mixed-precision, because BF16 has a wide dynamic range that is the same as FP32. LLM training and tuning with 8-bit low-precision were not well-explored in previous works, because the hardware support for FP8 is not available before the release of Nvidia Hopper infrastructure. This work presents the first exploration of FP8 pre-training and fine-tuning for LLMs, while proposing an extremely-optimized FP8 mixed-precision scheme. We hope this work could facilitate future research in FP8 and, potentially, extend to exploring even lower precision training, such as 4-bit and 1-bit.

\section{Conclusion}

In this work, we explore 8-bit training for LLMs. We introduce a new FP8 mixed-precision training framework, which incorporates 8-bit collective communication, optimizer, and distributed parallel training in an incremental manner. To our best knowledge, this is the first work infiltrating FP8 compute, storage and communication into the whole progress of large language model training. Extensive experiments demonstrate the proposed method effectively diminishes  communication overhead and curtails memory utilization in the context of GPT model training at various scales. %
In future work, we plan to scale up the size and training steps of the FP8 GPT models and further train them with our 8-bit mixed-precision scheme. Moreover, we will also use the proposed FP8 scheme to train multi-modal large  models, and explore low-bit deployment of LLMs on various edge devices, such as smart phones.

\vspace{3em}
\section*{Contribution and Acknowledgement}
\label{contribution}

This project was initially proposed by Han Hu and Peng Cheng, who are the directional lead. %
Shuguang Liu served as the product lead throughout the project. 

The contributions for all the co-authors are detailed as follows:

\textbf{FP8 Framework:}
Kan Wu, Houwen Peng, Ze Liu, Peng Cheng, Han Hu

\textbf{System:}
Yifan Xiong, Ziyue Yang, Yuxiang Yang, Guoshuai Zhao, Peng Cheng

\textbf{Hardware Infrastructure:}
Guoshuai Zhao, Yuxiang Yang, Yifan Xiong, Peng Cheng, Shuguang Liu, Joe Chau

\textbf{Data:}
Ruihang Li, Miaosen Zhang, Jia Ning, Chen Li, Ruizhe Wang, Houwen Peng, Han Hu

\textbf{Pre-training:}
Yixuan Wei, Kan Wu, Ze Liu, Miaosen Zhang, Zheng Zhang, Houwen Peng, Han Hu

\textbf{Alignment} (SFT, RS, and RLHF):
Bolin Ni, Jingcheng Hu, Yixuan Wei, Houwen Peng, Han Hu

\textbf{Evaluation:}
Yixuan Wei, Bolin Ni, Jingcheng Hu

\textbf{Product Engineering:}
Yuxiang Yang, Kan Wu, Yifan Xiong, Ziyue Yang, Guoshuai Zhao, Peng Cheng

\vspace{2em}

We thank Eric Chung, Bita Darvish Rouhani, Yu Pei, Hyunseung Harry Yoo, Zhenghong Zhou, Gongrui Zhang, and Zhirong Wu for helpful discussions.

We thank Baining Guo and Lidong Zhou for their guidance and support for this project.

\newpage
\bibliographystyle{plainnat} 
\bibliography{fp8-lm}

\clearpage
\appendix

\section{Appendix}

\subsection{FP8 Data Formats}
\label{appendix.A}

In September 2022, NVIDIA, ARM, and Intel published FP8 specification for standardization as an interchange format for AI \citep{fp8-dl}. The industry has moved from 32-bit precision to 16-bit, and now even 8-bit precision for AI model training. This development reflects a broader industry trend that has transitioned from high-precision to low-precision training. Notably, the proposed FP8 specification introduces two distinct data types, $E$5$M$2 and $E$4$M$3, which offer a trade-off between a larger range and higher precision of stored values \citep{te-fp8}.
\begin{itemize}
\item $E$4$M$3 consists of 1 sign bit, 4 exponent bits and 3 bits of mantissa. It can store values up to +/-448 and NaN.
\item $E$5$M$2 consists of 1 sign bit, 5 exponent bits and 2 bits of mantissa. It can store values up to +/-57344, +/- inf and NaN.
\end{itemize}

The FP8 format \citep{fp8-dl} roughly follows the IEEE 754 standard. Compared to higher precision data formats such as FP16 and FP32, FP8 suffers from two kinds of representation degradation:

\begin{itemize}
\item \emph{Lower representation range.} The representation range in a data format specifies the range between the maximum and minimum values that the format can accurately represent. There are two modes, a normal mode, which defines a regular range with relatively constant precision, and a subnormal mode, which extends the range to represent smaller values with lower precision. The normal range  primarily depends on the number of exponent ($E$) bits, with more $E$ bits resulting in a larger normal range. On the other hand, the subnormal range is primarily influenced by the number of mantissa ($M$) bits, where an increase in $M$ bits leads to a larger subnormal range. As illustrated in Tab. \ref{tab.fp8range}, the representation range of FP8 is notably narrower compared to that of FP16 and FP32, especially in the case of the $S$1$E$4$M$3 sub-format ($S$ denotes the sign bit). This discrepancy represents the primary challenge when employing FP8 for training large models.

\item \emph{Lower representation precision.} The limited number of mantissa ($M$ bits) leads to quantization representation errors. Due to the considerably fewer $M$ bits in FP8, the representation precision of FP8 is substantially lower than that of FP16, as depicted in Tab. \ref{tab.fp8range}. This challenge stands as another significant hurdle when considering the use of FP8 for training large models.
\end{itemize}

FP8 consists of two sub-formats: $S1E4M3$ and $S1E5M2$. The former offers a narrower representation range but higher precision, while the latter provides a larger range but lower precision. These two sub-formats give users the flexibility to strike a balance between their requirements for range and precision in model training. %

\begin{table} [!h]
\centering  
\caption{Representation range and error for different data formats}  
\resizebox{0.95\textwidth}{!}{%
\begin{tabular}{c|c|c|c|c|c}  
\hline  
Data format & \multicolumn{3}{c|}{Representation Range} & \multicolumn{2}{c}{Maximum Relative Error} \\  
\cline{2-6}  
& Max normal & Min normal & Min subnormal & Min - Max (normal) & Min $\sim$ Max (subnormal) \\  
\hline  
\makecell[c]{FP32 \\(S1E8M23)} & $3.40 \times 10^{38}$ & $1.18 \times 10^{-38}$ & $1.40 \times 10^{-45}$ & $1.19 \times 10^{-7} \sim 5.96 \times 10^{-8}$ & $5.00 \times 10^{-1} \sim 1.19 \times 10^{-7}$ \\
\hline  
\makecell[c]{FP16 \\(S1E5M10)} & $65,504$ & $6.10 \times 10^{-5}$ & $5.96 \times 10^{-8}$ & $9.76 \times 10^{-4} \sim 4.89 \times 10^{-4}$ & $5.00 \times 10^{-1} \sim 9.78 \times 10^{-4}$ \\
\hline  
\makecell[c]{BF16 \\(S1E8M7)} & $3.39 \times 10^{38}$ & $1.18 \times 10^{-38}$ & $9.18 \times 10^{-41}$ & $7.75 \times 10^{-3} \sim 3.94 \times 10^{-3}$ & $5.00 \times 10^{-1} \sim 7.94 \times 10^{-3}$ \\
\hline  
\makecell[c]{FP8 \\(S1E4M3)} & $448$ & $1.56 \times 10^{-2}$ & $1.95 \times 10^{-3}$ & $1.11 \times 10^{-1} \sim 7.69 \times 10^{-2}$ & $5.00 \times 10^{-1} \sim 1.67 \times 10^{-1}$ \\
\hline  
\makecell[c]{FP8 \\(S1E5M2)} & $57,344$ & $6.10 \times 10^{-5}$ & $1.53 \times 10^{-5}$ & $2.00 \times 10^{-1} \sim 1.67 \times 10^{-1}$ & $5.00 \times 10^{-1} \sim 5.00 \times 10^{-1}$ \\
\hline  
\end{tabular}  
}
\label{tab.fp8range}
\end{table}

\subsection{FP8 Tensor Scaling}
\label{appendix.B}

\begin{figure}[!h]
\centering
\scalebox{0.6}{
\includegraphics{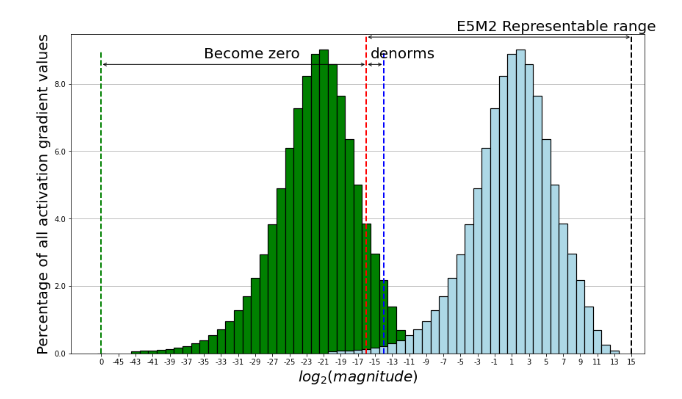}
}
\caption{Scaling gradients to fall within the representation range of the FP8 datatype.}
\label{fig:tensorscaling}
\end{figure}

We now discuss the underlying mechanisms for how large model training with FP8 overcomes the challenges associated with representation range and precision degradation. The key technique behind is tensor scaling, which scales the tensor values that originally locate out the representation range of a data format to its comfort zone, as visualized in Fig. \ref{fig:tensorscaling}. 
The pioneer scaling techniques \citep{amp,apex} apply a global scaling factor to the loss, such that gradients of all layers are scaled by a single adaptive factor. The utilization of the global loss scaling technique, in conjunction with various other training strategies, has facilitated the widespread adoption of FP16 mixed-precision training on V100 and A100 GPUs. Remarkably, this approach has resulted in minimal to no degradation in accuracy, particularly for small to medium-sized models \citep{amp}. Nonetheless, when dealing with super-large models or complex tasks, such as in the training of models like DALL-E \citep{dalle}, the global loss scaling technique still encounters significant underflow issues. As a consequence, block-wise \citep{dalle} and layer-wise \citep{4bit} gradient scaling are proposed. %

While the global scaling technique enables almost no accuracy drop for FP16 training (with a range of [5.96E-8, 6.55E+4]), the fine-grained per-tensor scaling will enable stable model training using even shallower range by FP8 (with a range of [1.95E-3, 448] for $E$4$M$3 and a range of [1.53E-5, 5.73E+4] for $E$5$M$2). Fig. \ref{fig:tensorscaling} shows that the representation range of FP8 has been large enough to deal with general model training.
In the per-tensor scaling technique, various strategies are available for choosing the suitable scaling factor for a given FP8 tensor. Two common approaches are ``just-in-time scaling" and ``delayed scaling" \citep{te-fp8}.

\begin{itemize}
\item \emph{Just-in-time scaling}. This strategy involves determining the scaling factor based on the maximum absolute value (amax) of the tensor being generated. However, in practical applications, this approach is often infeasible because it necessitates multiple passes through the data. Specifically, the operator first produces and writes out the output in higher precision, then calculates the maximum absolute value of the output, and finally applies this scaling factor to all values to obtain the final FP8 output. This process introduces a significant amount of overhead, which can substantially reduce the benefits of using FP8.

\item \emph{Delayed scaling}. This strategy involves selecting the scaling factor based on the maximum absolute values observed in a certain number of preceding iterations. This approach allows for the full performance benefits of FP8 computation but necessitates the storage of a history of maximum values as additional parameters of the FP8 operators.
\end{itemize}

\subsection{Pre-training Data}
\label{appendix.data}

Tab. \ref{tab.data} presents an overview of our collected data sources along with the corresponding sampling weights employed in pre-training. The arXiv and StackExchange subsets are collected from Redpajama \citep{redpajama}, while BookCorpus2 \citep{bookcorpus}, Books3 \citep{Presser2020Books3}, DM-Math \citep{dm-math}, Gutenberg \citep{pg19}, HackerNews\footnote{https://news.ycombinator.com}, NIH ExPorter\footnote{https://exporter.nih.gov}, OpenSubtitles \citep{opensubtitles}, and USPTO\footnote{https://bulkdata.uspto.gov} subsets are extracted from The Pile \citep{pile}. The Wikipedia data is downloaded from HuggingFace \citep{wikihf}. We use the 20220301 dump, including 24 languages: bg, ca, cs, da, de, en, es, fr, hi, hr, hu, it, jp, ko, nl, pl, pt, ro, ru, sl, sr, sv, uk, zh. %

We pre-process 11 CommonCrawl snapshots, ranging from 2018 to 2023, with the CCNet pipeline \citep{wenzek2019ccnet}. This process involves data deduplication at the line level, followed by language identification utilizing a fastText linear classifier \citep{fasttext} to eliminate non-English pages. A filtering mechanism based on an n-gram language model is employed to exclude low-quality content. In addition, we train a linear classifier \citep{redpajama} to distinguish documents similar to Wikipedia pages from randomly sampled CommonCrawl documents. Documents not classified as resembling Wikipedia are excluded. Finally, we perform fuzzy deduplication \citep{lee_deduplicating_2022} across all the processed snapshots from CommonCrawl.

We collect Python code data from Github using a repository list provided by Bing indexing \citep{bing}. The cleaning of the code data includes three steps. First, we remove control characters, except for $\textbackslash t$ and $\textbackslash n$. Next, we remove copyright comments in the code. An alphanumeric rate filter is then applied, removing lines with a rate below 0.5 if they are comments, and discarding the entire file if its overall alphanumeric rate is less than 0.98. Files with less than 5 lines or a maximum line length exceeding 1,000 characters are also discarded. Also, files with an average line length of more than 100 characters are discarded. Lastly, a pattern search is conducted to identify key Python keywords (\emph{e.g.}, import, from, def, class, if, for, try, etc.) within the code. Files containing less than 3 instances of these keywords are eliminated. This comprehensive process ensures that the remaining Python code data is of high quality and suitable for use in academic research. We additionally add Python code from Stack \citep{kocetkov2022stack}, and perform fuzzy deduplication within all the collected Python code.
\begin{table}[h!]
\centering  
\scalebox{1.0}{
\begin{tabular}{c|c|c|c}  
\toprule  
\textbf{Dataset} & \textbf{Sampling prop.} & \textbf{Epochs} & \textbf{Training Tokens (Billion)} \\  
\midrule  
\multicolumn{4}{c}{\textbf{Web Crawls}} \\  
\midrule  
CommonCrawl & 51.71\% & 0.16 & 51.71 \\  
C4 & 25.56\% & 0.16 & 25.56 \\  
OpenWebText & 2.73\% & 0.16 & 2.73 \\  
\midrule  
\multicolumn{4}{c}{\textbf{Technical \& Science content}} \\  
\midrule  
arXiv & 1.54\% & 0.05 & 1.54 \\  
StackExchange & 1.42\% & 0.08 & 1.42 \\  
DM-Math & 0.39\% & 0.05 & 0.39 \\  
USPTO & 0.52\% & 0.05 & 0.52 \\  
NIH ExPorter & 0.04\% & 0.05 & 0.04 \\  
\midrule  
\multicolumn{4}{c}{\textbf{Programming Languages}} \\  
\midrule  
Python & 4.50\% & 0.11 & 4.50 \\  
\midrule  
\multicolumn{4}{c}{\textbf{Other Curated Sources}} \\  
\midrule  
Wikipedia & 4.50\% & 0.16 & 4.50 \\  
Books & 4.50\% & 0.09 & 4.50 \\  
News & 2.00\% & 0.11 & 2.00 \\  
Dialogue & 2.00\% & 0.27 & 2.00 \\  
\midrule  
\multicolumn{3}{c|}{\textbf{Total}} & 100.00 \\  
\bottomrule  
\end{tabular}  
}
\caption{Pre-training data. For each subset we list the sampling weight, number of epochs, and training tokens. Books data includes BookCorpus2 \citep{bookcorpus}, Books3 \citep{Presser2020Books3}, and Gutenberg \citep{pg19}. Dialogue data includes HackerNews and OpenSubtitles \citep{opensubtitles}. For experiments with a training token count of less than 100 billion, we employ the same sampling proportion. }  
\label{tab.data}
\end{table}
\end{document}